\documentclass[10pt,journal,compsoc]{IEEEtran}
\usepackage{graphicx}
\usepackage{comment}
\usepackage{amsmath,amssymb} 
\usepackage{color}
\usepackage{wrapfig}

\definecolor{commentRonnie}{rgb}{0.9, 0.1, 0.1}

\newcommand{\etal}{\textit{et al.}}
\ifCLASSOPTIONcompsoc
  \usepackage[nocompress]{cite}
\else
  \usepackage{cite}
\fi
\ifCLASSINFOpdf
\else
\fi
\begin{document}
%
\title{Orientation Keypoints for 6D Human Pose Estimation}
%
%
%
%

\author{Martin~Fisch,
        Ronald~Clark
\IEEEcompsocitemizethanks{\IEEEcompsocthanksitem Department of Computing, Imperial College London.\protect\\
}
\thanks{Manuscript received July 10, 2020.}}

%
%

\markboth{Orientation Keypoints for 6D Human Pose Estimation}%
{Shell \MakeLowercase{\textit{et al.}}: Bare Demo of IEEEtran.cls for Computer Society Journals}
%



\IEEEtitleabstractindextext{%
\begin{abstract}
  Most realtime human pose estimation approaches are based on detecting joint positions. Using the detected joint positions, the yaw and pitch of the limbs can be computed. However, the roll along the limb, which is critical for application such as sports analysis and computer animation, cannot be computed as this axis of rotation remains unobserved. In this paper we therefore introduce orientation keypoints, a novel approach for estimating the full position and rotation of skeletal joints, using only single-frame RGB images.  Inspired by how motion-capture systems use a set of point markers to estimate full bone rotations, our method uses virtual markers to generate sufficient information to accurately infer rotations with simple post processing. The rotation predictions improve upon the best reported mean error for joint angles by 48\% and achieves 93\% accuracy across 15 bone rotations.  The method also improves the current {\color{black} state-of-the-art} results for joint positions by 14\% as measured by MPJPE on the principle dataset, and generalizes well to in-the-wild datasets. 
\end{abstract}

}

\maketitle

\IEEEdisplaynontitleabstractindextext

%
\IEEEpeerreviewmaketitle


\IEEEraisesectionheading{\section{Introduction}\label{sec:introduction}}

Human motion capture (MoCap) has been a major enabling technology across both the arts and sciences. Motion capture has played a key role in kinematic analysis for sports and medicine, has created engaging user experiences with devices like the Kinect, and has been an essential part of the visual effects industry for years. In the past, MoCap required sophisticated purpose-built studios with multi-camera capture systems. However, recent advances in computer vision have led to new ways of doing MoCap that are far less restrictive than traditional methods. These approaches can capture 3D human poses from single RGB cameras and have spurred interest in next-generation applications such as personal digital sports coaches, and the possibility of capturing high-quality animation directly on consumer smartphones. However, despite their great promise, existing single-camera human pose estimation approaches have failed to achieve a level of fidelity that matches that of traditional MoCap systems.

\begin{figure}[t!]
\centering
\includegraphics[width=\linewidth]{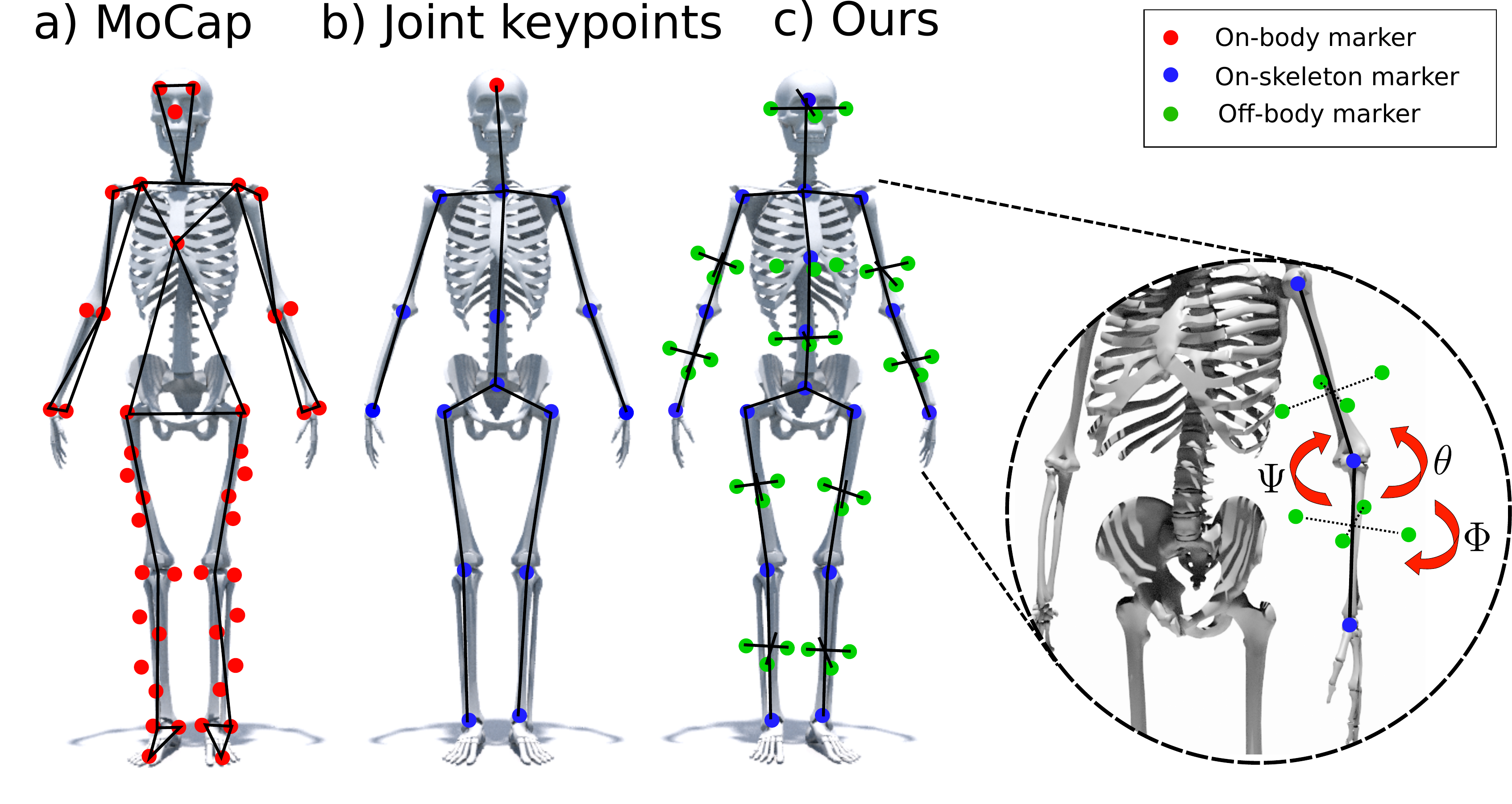}
\caption{Our orientation keypoints approach uses a set of virtual point markers, similar to that used in motion capture systems, to accurately infer joint positions and orientations. a) Motion capture systems place sufficiently many (physical) markers to fully observe rigid joint rotations, including roll $\Phi$, yaw $\Psi$ and pitch $\theta$ of each bone. b) Existing joint keypoint-based human pose estimation methods place keypoints at joints, which only allows $\Psi$ and $\theta$ to be estimated, leaving roll unobserved. c) Our approach estimates virtual markers or "orientation keypoints" capturing the full joint rotation. }
\label{fig:OKPS}
\end{figure}

Much of the current research in 3D pose estimation focuses on localizing joint keypoints with convolutional neural networks (CNNs). However, as shown in Figure \ref{fig:OKPS} (b), most methods only detect keypoints at the joint locations. Using these detected points, the yaw, $\Psi$, and pitch, $\theta$, can be computed however, one degree of freedom is left unobserved, i.e., the roll, $\Phi$, around the axis. Therefore, most keypoint-based human pose estimation approaches can only observe five degrees-of-freedom for each joint, although there are six degrees of freedom, i.e., (x, y, z, $\Phi$, $\Psi$, $\theta$). 

In order to address this problem and estimate the full six the degrees of freedom, we propose a method that takes inspiration from traditional MoCap systems. MoCap systems use a large set of markers attached to the body, as shown in Figure \ref{fig:OKPS} (a). Groups of these markers are used to compute the orientation of each bone. For example, the upper leg has four markers attached, which are used to solve the femur's position and orientation. 

Inspired by this, we introduce a novel keypoint-based approach that solves for complete kinematic transforms of a human skeleton, with six degrees of freedom at each bone as illustrated in Figure \ref{fig:OKPS} (c).  Key to our approach is an additional set of \textit{orientation keypoints} that provide sufficient information for inferring full joint orientation. 
\vspace{5mm}

\noindent Specifically, our contributions in this paper are threefold:
\begin{enumerate}
\item We introduce orientation keypoints as a novel approach to solving for full 3-axis joint rotations.
\item We propose a neural network model that accurately localizes these points in 3D from monocular images, achieving {\color{black} {\color{black} state-of-the-art}} accuracy for joint rotation and position estimation.  
\item We demonstrate that our approach generalizes well to other datasets even without retraining, and with fine-tuning we set {\color{black} state-of-the-art} benchmarks on these additional sets as well.  
\end{enumerate}

\section{Related Work}

Deep learning related pose estimation initially focused on estimating 2D poses, with many of these techniques subsequently extended to 3D.  For relevance, we focus on monocular single-image pose estimation works here.  

\textbf{Pose Estimation in 2D}.  Many approaches \cite{Tompson_2015_CVPR,Xiao_2018_ECCV,sun2018integral} predict discrete heatmaps estimating the probability of a joint occurring at each pixel instead of a continuous regression. Converting between the heatmaps and point coordinates can be done in several ways. The most common is to either take a ``hard'' argmax of the maps, or via integral regression (i.e., a ``soft'' argmax) \cite{sun2018integral}.  We use soft argmax for this paper as it delivered slightly more accurate quantitative results.  The architecture of these approaches vary, \cite{Xiao_2018_ECCV} use a ResNet backbone and three convolution transpose layers to upsample into 64x64 pixel heatmaps while \cite{Newell-StackedHourglass} introduce the ``stacked hourglass'' architecture and \cite{He_2017_ICCV} use Mask-RCNN and pixel-by-pixel masks to predict keypoints. One of the most successful approaches has been the cascaded pyramid network (CPN) which aims to address the problem of hard keypoints, integrate feature representations and use online hard keypoint mining loss (OHKM)  \cite{Chen_2018_CVPR}.   Augmenting the keypoint heatmaps with part affinity fields has also shown to be very beneficial especially when predicting poses for a variable number of people \cite{cao2018openpose}. In contrast to these works, our approach operates in 3D but can also be used to predict 2D keypoints simply by dropping the z-dimension in the predictions.

\textbf{Pose Estimation in 3D}
Multiple 3D interpretations typically exist for a single 2D skeleton \cite{Lee1985DeterminationO3}. Therefore many approaches  use a preconstructed model to map 2D detections to 3D \cite{Akhter_2015_CVPR,chen20173d,Jiang20103DHP,Tome_2017_CVPR,yasin2016dual}. The first approaches along these lines created a pose dictionary from 3D MoCap data to generate paired 2D projections from different angles and generate depth values with a lookup \cite{Akhter_2015_CVPR}, while others have used a nearest neighbor search \cite{chen20173d,Jiang20103DHP}. Geometric information, such as bone length priors and projection consistency, can also be utilized for converting from 2D to 3D \cite{brau20163d}. Other approaches take this a step further by trying to establish the correspondence between the 2D image and a 3D human model.  This has been done, for example, by using distance matrix regression \cite{Moreno-Noguer_2017_CVPR} or by directly predicting dense correspondences between pixels in the image to UV coordinates on a body mesh \cite{Pons2015} or landmark locations on the body  \cite{lassner_unite}. This differs from our research because we do not use body landmarks or a body model but instead chose detached points to maximize the angular perspective. 

Another popular paradigm is to ``lift'' 2D detections to 3D using a learned network \cite{Tome_2017_CVPR,Moreno-Noguer_2017_CVPR,martinez_2017_3dbaseline,Pons2015}. In fact, \cite{martinez_2017_3dbaseline} showed that lifting ground truth 2D locations to 3D can be solved with a low error rate with a relatively simple network.  Other approaches perform a direct 3D prediction of keypoints from the images \cite{pavlakos2017coarse-to-fine,Park20163DHP,tekin2017learning,Habibie_2019_CVPR}. Many of these approaches are voxel-based which can be memory intensive and requires discretizing the space at a suitable resolution. To overcome this, \cite{pavlakos2017coarse-to-fine} propose a fine discretization of the 3D space around the subject and train a network to predict per voxel likelihoods for each joint. Combining 2D information with the 3D predictions can also help improve accuracy, and therefore \cite{Park20163DHP,tekin2017learning} fuse direct image 3D features with 2D estimation while      \cite{Habibie_2019_CVPR} embed 3D pose cues in a learned latent space.  
In this paper we consider two models, one which predicts 3D keypoints through regression and one which predicts through per dimension heatmaps.

\textbf{Weak Supervision and Generative Approaches.}
As obtaining ground-truth labels for keypoints can be challenging, several works have focussed on using other signals for training.  Geometric constraints can be used to train on in-the-wild datasets in a self-supervised manner \cite{Zhou_2017_ICCV,kocabas2019epipolar,iqbal2020weakly, drover2018can}.  Other, weaker supervision signals can also be used such the ordinal depths of human joints, acquired from supplementary human annotations \cite{pavlakos2018ordinal}. Adversarial training has also been quite popular as it enables using unlabelled data for training or training 3D predictions with only 2D annotations \cite{wandt2019repnet}. This is usually accomplished by generating 3D pose predictions for images with only 2D annotations and using a discriminator which distinguishes implausible poses \cite{Yang18,Chen_2019_CVPR,kanazawaHMR18,wandt2019repnet}. In our approach we do not use weak supervision, as some works have reported convergerce issues with GAN-type losses, however, this could easily be included in our framework in the future. 

\textbf{Estimation from video.}
While most works have focussed on the single-frame setting, utilizing the temporal regularity of video can help to improve pose accuracy. The temporal regularity can be integreted in various ways. Some approaches \cite{Zhou-MonoCap} use explicit temporal smoothness constraints, while others have used  recurrent LSTM / GRU units \cite{rayat2018exploiting,fragkiadaki2015recurrent, kocabas2020vibe}, and dilated temporal convolutions  \cite{DBLP:PavLLO_3Dhumanposeinvideo}.  While our approach only relies on single frames, the technique can easily be extended to most multi-frame settings.

\textbf{Joint rotation prediction.}
Existing research which directly estimates joint angles, can capture the full six degrees of freedom when used in conjunction with kinematic constraints of a skeleton model. Here, 3D joint positions are typically computed by using for the forward kinematics. However, these approaches significantly underperform location-based methods, as convolutional neural networks have not proven adept at modeling the non-linearities complexities of angular representations.

Various parameterizations can be used for the joint angles, such as quaternions \cite{pavllo:quaternet:2018}, Euler angles \cite{XNect} or by regressing 3x3 rotation matrices \cite{yoshiyasu_accv_2018}. The estimated joint angles are then usually mapped onto a skeleton using forward kinematics \cite{XNect}. The angular constraints within the skeleton's kinematic chain can be formulated in a differentiable manner and embedded directly into the network itself \cite{zhou_eccv_16}. There is also a body of related work which predict 6D position and rotation of objects by estimating virtual 3D bounding box vertices \cite{Rad_2017_ICCV,Tekin_2018_CVPR}. This is conceptually the closest to our approach, but we calculate 15 rotations of a highly complex kinematic chain using a mix of joint and virtual markers detached from the shape of the limb. 

\textbf{Direct mesh regression}

{\color{black} There are two types of methods in this line of work, the first type of approach does not regress mesh vertices / correspondences directly but use the “Skinned Multi-Person Linear” (SMPL) model \cite{SMPL:2015}, or the newer SMPL-X \cite{SMPL-X:2019}, which represent the mesh as a set of low dimensional parameters. In particular, SMPL represents the mesh as 10 shape parameters and 72 (24*3) pose parameters. The pose is specified as relative rotation vectors which define the angle between successive joints. For pose estimation, SMPL-based methods either regress the shape and joint angles directly from an image \cite{choutas2020monocular}, or from various types of intermediate representation, including 2D keypoints and silhouette \cite{pavlakos2018humanshape} or semantic segmentation of the body parts \cite{omran2018nbf}. Other works do not regress the parameters directly, but predict more easily observable quantities such as joint positions and use an optimization to find the joint angles and shape parameters\cite{bogo_smplify}. In this sense, our method is complementary to SMPL in that the joint angles estimated using our OKPs can be used to drive the SMPL model. However, unlike general SMPL approaches, our method does not require mesh annotations for training. 

The second type of method directly regresses correspondences between a dense set of points in the RGB image and uv-locations on the canonical human mesh. This dense set of points lies on the mesh and the network only predicts those which are not occluded. In order to recover body pose (i.e. joint position and angles), these methods typically also make use of SMPL \cite{zhang2020learning}. Specifically, they first regress dense uv-coordinates, followed by an estimation of the SMPL parameters conditioned on this information. As these methods only predict correspondences for visible (non-occluded) points in the image, estimating the pose of the full skeleton is difficult, and requires a complex setup such as that in \cite{zhang2020learning} to obtain good performance. Furthermore, these methods require training data where dense correspondences have been established between the image pixels and mesh vertices. In general, this data is very difficult to obtain. 

}


\textbf{Deep rotation estimation.} As rotations play an important role in many tasks, there has been a growing theoretical interest in finding out what representations work best with deep networks. The fundamental issue with representing rotations is that lie in the special orthogonal group, $SO(3)$, which consists of all orthogonal $3\times3$ matrices. This constraint means that any parameterization using less than 5 dimension is guaranteed to be discontinuous, which creates problems when training deep networks \cite{Zhou_2019_CVPR}. To address this, \cite{Zhou_2019_CVPR} propose a continuous representation for rotations using a 6D over-parameterization and the Gram-Schmidt procedure to recover the rotation matrix. More recently, in concurrent work, \cite{levinson2020analysis} showed that using a 9D over-parameterization followed by SVD to recover the rotation outperforms the Gram-Schmidt procedure of \cite{Zhou_2019_CVPR} in terms of accuracy. In contrast, we introduce a representation for bone poses in $SE(3)$, including both the translation and rotation, using a 12D over-parameterization called orientation keypoints. We are also the first to apply this concept of rotation over-parameterization to human pose estimation.

\section{Approach}

In this section we describe our approach for estimating human pose. In Section \ref{sec:okps} we present orientation keypoints (OKPS), the main component of our approach which is used as an intermediate representation from which bone rotations and translations can be computed. We then describe the networks we use to predict our OKPS and introduce our novel crosshairs architecture in Section \ref{sec:networks}. Finally, we describe how we post-process the OKPS detections to obtain bone translations and rotations.

\begin{figure}[t!]
    \centering
    \includegraphics[width=\columnwidth]{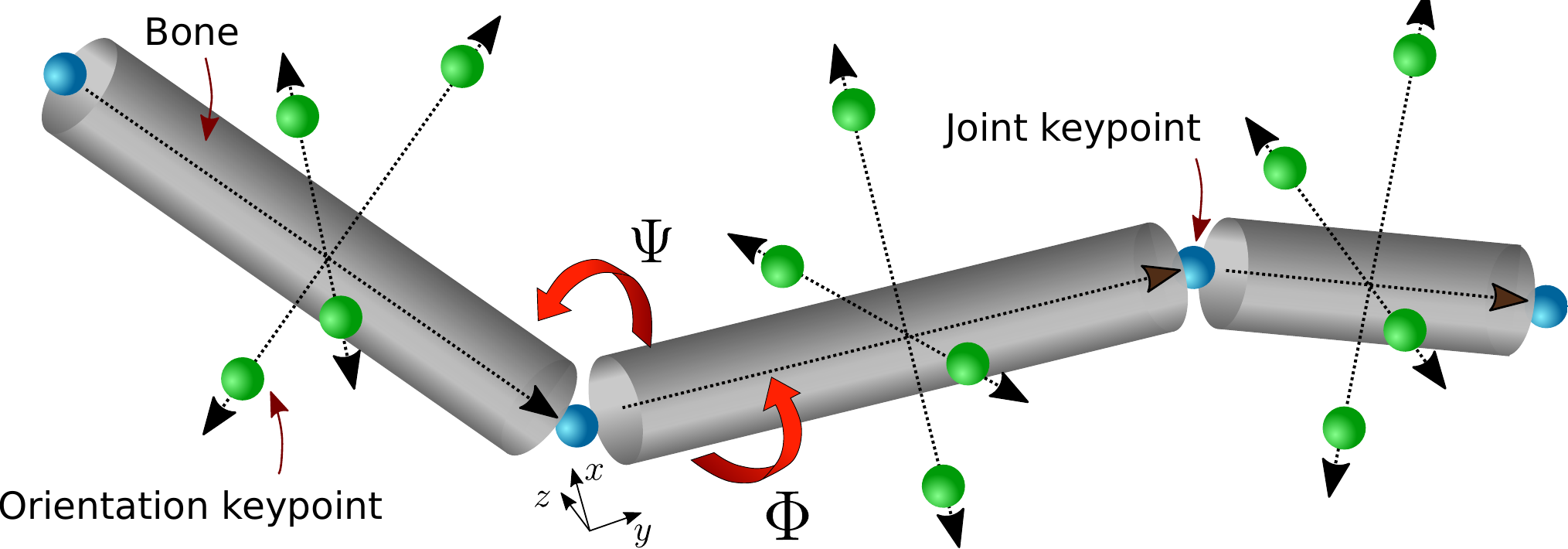}
    \caption{Illustration of the arrangement of the joint keypoints (blue) and orientation keypoints (green) with respect to their parent bones. Without the orientaiton keypoints, the roll $\Phi$ is not observable.}
    \label{fig:OKPS_illustration}
\end{figure}

\subsection{Orientation keypoints}
\label{sec:okps}

Conventional keypoint-based approaches for 3D human pose estimation focus on predicting a set of joint keypoints (JKPS), 

\begin{equation}
    P_{jkps} =\left(\mathbf{x}_{1}, \mathbf{x}_{2}, \mathbf{x}_{3}, \ldots, \mathbf{x}_{{N_j}}\right)
\end{equation}  
where each $p_{j_i}=(x, y, z)^{\top}$ describes the 3D location of a specific joint. The number of JKPS, $N_j$ is typically around 17, corresponding to joints on the human body such as the elbows, ankles, knees, wrists, hip, neck, etc. However, as alluded to in the introduction, this set of keypoints does not capture all the degrees of freedom of the bones of the human body. Therefore, we introduce an extra set of ``orientation keypoints'' (OKPS),

\begin{equation}
    P_{okps} =\left(\mathbf{x}_{N_j}, \mathbf{x}_{N_j+1}, \mathbf{x}_{N_j+2}, \ldots, \mathbf{x}_{{N_j + N_o}}\right),
\end{equation}  

and combine the two sets $P = P_{jkps} \bigcup P_{okps}$ for the purpose of pose estimation. 

We define the OKPS as clusters of points rigidly attached to a particular joint keypoint to provide information about the two axes ignored by conventional JKPS.  They differ from dense pose correspondences and landmarks in that they do not directly correspond to a specific body part or shape but are instead anchored in specific directions from the center of the bone (i.e., forward, or to the side) well offset from the body. These are inspired by ``real'' MoCap markers which are usually retro-reflective white balls that are attached rigidly to the actor. Orientation keypoints are therefore analogous to MoCap markers, but with the major advantage that no actual marker needs to be attached to the actor -- they are simply virtual keypoints detected in relation to natural landmarks on the body. The difference is demonstrated in Figure \ref{fig:OKPS}.  For example, we assign an orientation keypoint for the lower-left leg set midway between the knee and ankle and well offset from the shin by half the leg bone's length.  In this paper, we use four OKPS for each of the 15 free rotations in the 17 joint skeleton.  Each OKPS is rigidly attached to the corresponding parent joint at a distance scaled by the bone length, one forward, back, left, and right, defined in the neutral pose. In general, given a bone $k$ with its first joint keypoint, $x_{j_1}$, the four OKPS for the bone are defined as:

\begin{align}
    \mathbf{x}_{o_1} &= \mathbf{x}_{j_1} + T_k \times l_k(0.5, 0.5,0)^\top \nonumber \\ 
    \mathbf{x}_{o_2} &= \mathbf{x}_{j_1} + T_k \times l_k(0.5,-0.5,0)^\top \nonumber \\ 
    \mathbf{x}_{o_3} &= \mathbf{x}_{j_1} + T_k \times l_k(0.5, 0,  0.5)^\top  \nonumber \\ 
    \mathbf{x}_{o_4} &= \mathbf{x}_{j_1} + T_k \times l_k(0.5, 0, -0.5)^\top
\end{align}

where $l_k$ is the length of bone $k$, and $T_k$ is the transformation (i.e. pose) that converts coordinates in the bone's keyframe to world space.


\subsection{Network design}
\label{sec:networks}

Our framework's main component is a convolutional neural network detector that localizes both joint and orientation keypoints.  Since the latter is offset and virtual, this requires learning depth and perspective even for 2D predictions, and so we choose to predict the full 3D keypoint locations directly in our detector.  This also enables direct calculation of the full kinematic rotations from the model without further lifting.  We also explore using a two-stage process similar to \cite{martinez_2017_3dbaseline} and \cite{DBLP:PavLLO_3Dhumanposeinvideo}, where the second stage is a lifter model which transforms 2D detections into 3D predictions (without a detector depth branch), or a refiner model which further hones initial 3D predictions.  

\subsubsection{Detector models}

For the detector, we experiment with two models; a simple Resnet based 3D regression baseline and a more sophisticated novel architecture, which we call Crosshairs, capable of providing accurate 3D estimates while limiting the memory and calculation overhead.  We describe them both below.

\textbf{Simple regression baseline}
This simple baseline detector uses a Resnet50 as the backbone and adds a head connected to the final convolution layer (removing the final pooling and fully connected layer in the base Resnet).  Our simple head is composed of four layers using grouped convolutions.  The grouping focuses the model at each keypoint and considerably limits the calculations beyond the backbone. The overall architecture consists of,

\begin{itemize}
\item  \textit{A convolutional layer} with a 1x1 kernel, batchnorm and ReLU.  We use 12 x Number of Keypoints (77) = 924 channels.  This prepares features for the grouped-by-keypoint convolutions
\item  \textit{A second grouped convolutional layer} with a 5x5 kernel (2x2 padding), batchnorm and ReLU.  We use the same number of channels but with 77 groups.  This means for each keypoint there are 12 convolutional filters each only using 12 channels from the previous layer. 
\item  \textit{A grouped convolutional layer} with a 1x1 kernel, batchnorm and ReLU.  The 924 channels and 77 groups are the same.
\item  A final \textit{fully connected grouped convolutional layer}.  As the kernel is the full width and height of the layer (12x9) this acts as a fully connected layer, except it is again grouped by keypoint: each keypoint's xyz prediction is made only from the 12 associated channels.  The number of channels is 3 (for XYZ) x Number of Keypoints, each outputting a single value.  
\end{itemize}

This detector is simple, very fast, and quick to train. In terms of accuracy, it is a strong baseline model and we show good results for Orientation Keypoints, even using this baseline detector. As we show in Table \ref{tab:ablation1}, it is sufficient to achieve {\color{black} state-of-the-art} results in 3D on Human 3.6m when used with orientation keypoints.

\textbf{The Crosshairs detector}
Conceptually we follow \cite{Xiao_2018_ECCV} and use a Resnet backbone with convolution transpose layers to recover a higher resolution.  Our key innovation is that at the output, each strand uses 1D heatmaps per dimension in place of square and volumetric heatmaps for 2D and 3D estimation.  This keeps the computation cost of the head linear with the resolution rather than square or cubic and is a substantial saving, particularly considering the larger number of keypoints we employ.

\begin{figure*}[]
  \centering
  \includegraphics[width=0.8\textwidth]{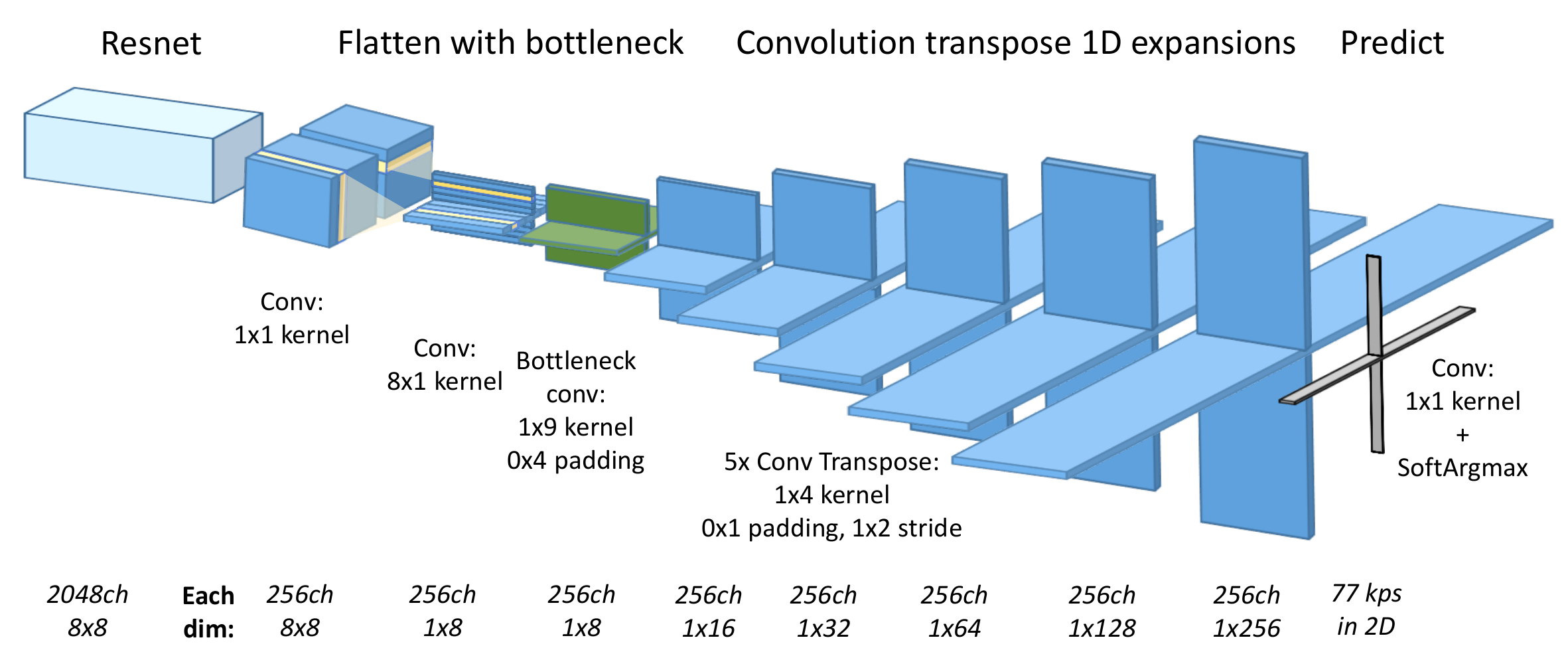}
  \caption{Overview of the crosshairs detector head.  We visualize only a 2D detector with single strands for X and Y attached to the C5 Resnet layer.  Each strand flattens the backbone layer along one dimension, filters through a bottleneck with a full width kernel, and then expands the single dimension back to the original resolution with convolution transpose layers.  Batch normalization and ReLU activation layers follow convolutions.  A 1x1 convolution layer and soft argmax make predictions.}
\label{Fig: architecture}
\end{figure*}

For example, one \textit{x} strand takes the C5 2048 channel 8x8 tensor and first samples into a 256 channel 8x8 tensor using 1x1 kernel convolutions. The y-dimension is then flattened into a tensor with only the x-dimension using an 8x1 convolution.  Flattening is followed by a bottleneck block with padding and a kernel size of 1x9 to immediately provide a global view - in a single dimension and at a low resolution, which is computationally efficient.  We then use transpose convolution layers to upsample, in one-dimension, back to the original resolution width.  The upsampling approach is similar to \cite{Xiao_2018_ECCV} but much cheaper, as we are operating in a single dimension.  Each convolution and convolution transpose layer is followed by a batchnorm layer \cite{batchnorm} and ReLU \cite{relu} activation function.  We use 256 channels throughout. A final 1x1 convolution layer collapses the channels into a 1D heatmap for each keypoint along the single dimension.  Each heatmap represents the network's estimate of the keypoint position along the single axis. We then apply the same technique for the Y dimension by instead flattening the other dimension.  As depth is not a native dimension, we use the same principle but a modified flattening procedure.  Again, we sample the backbone C5 layer with 1x1 convolutions but with more channels to reshape into a depth dimension, i.e. into a 256 channel 9x9x12 tensor (we arbitrarily decide the tensor depth resolution is equal to the narrowest XY dimension).  We then use a convolution layer and a 1x9x12 kernel, which only slides in the depth dimension, to collapse x and y into a 9x1x1 block, which is again a flattened to 1D as before.  

This leads to 3 vectors for each point representing the same data as a volume would but much more efficiently,
\begin{equation}
    \mathbf{v}^x = (v^x_k)_{k=1}^{N_x}, \mathbf{v}^y = (v^y_k)_{k=1}^{N_y}, \mathbf{v}^z = (v^z_k)_{k=1}^{N_z}
\end{equation}

The keypoints are then recovered by using a coordinate-weighted softmax \cite{Luvizon_2018_CVPR}: 

\begin{equation}
   \mathbf{x}_k = \left( \frac{\sum_i e^{v^x_{i}} w^x_{i}}{\sum_i e^{v^x_i}}, \frac{\sum_i e^{v^y_{i}} w^y_{i}}{\sum_i e^{v^y_i}}, \frac{\sum_i e^{v^z_{i}} w^z_{i}}{\sum_i e^{v^z_i}} \right)
\end{equation}

where $w^x_i = \frac{i-0.5 N_{x}}{0.5 N_{x}}$ so that $w_i \in [-1,1]$.

The introduction of orientation keypoints adds points that may lie outside the subject's silhouette and a tighter fitting bounding box.  Rather than use a larger bounding box and lose effective resolution, we instead map the soft argmax layer output to a 25\% wider pixel range than the underlying image - thus each heatmap covers a wider area than the image itself. 

For the benefits of intermediate supervision and higher resolution access, we propose using multiple crosshairs, one attached to each layer group of Resnet.  We aggregate crosshair strands with a 1x1 kernel convolution layer combining the concatenated high-resolution 1D feature maps, followed by batchnorm, ReLU, and a bottleneck block with a 1x5 kernel. This produces the final predictions.

\subsubsection{Lifter/refiner regression model}
\label{sec:refiner}

The lifter takes as input the set of normalized keypoints predicted by the detector (either a 77x3 or 77x2 matrix depending on whether a 2D or 3D detector is used) and outputs the keypoints in metric space. For the architecture of the lifter/refiner block we follow \cite{martinez_2017_3dbaseline} and use a similar architecture.  This entails an inner block with a linear layer, followed by batch normalization \cite{batchnorm}, dropout \cite{dropout} and rectified linear units \cite{relu}.  The outer blocks contain two inner blocks and a residual connection.  A first linear layer converts from the number of keypoint inputs, flattened to a single dimension, into the network's linear width.  A final layer converts from the width to the number of predictions. We use two outer blocks in our lifter/refiner, but widen the network compared to \cite{martinez_2017_3dbaseline}, increasing the size of each linear layer by 50\% from 1024 to 1536, which approximately doubles the total parameters.  This helps accommodate the 5-6x as many keypoint inputs and outputs needed for orientation keypoints.

\subsection{Post processing}
\label{sec:post_processing}

For inference, we take the average of the predictions and the horizontally flipped predictions and use these to compute the rotations and positions of each joint.

\textbf{Positions.} As the detector predicts keypoints in normalized pixel or voxel units, we consider two approaches to predicting real-world 3d positions from orientation keypoints : (a) we map the rotations onto a full kinematic skeleton, based on the average bone lengths from the training set, and (b) train a second stage refiner network described in Section \ref{sec:refiner}.  This can lift purely from 2D or refine 3D voxels.  The first approach more elegantly unites rotations with positions for 6D, easily allows remapping onto different sized individuals in new environments and can deal well with novel poses.  Differences in skeleton size however, contribute to the error.  The second approach effectively bakes the skeleton size and camera perspectives into an additional neural network during training and is more reliant on the set of training poses, as highlighted in \cite{martinez_2017_3dbaseline}, but is more accurate in the Human 3.6m setting.

\textbf{Rotations.} We calculate each joint's rotation from the two JKPS and four OKPS associated with each bone with reference to the neutral T-pose positions of these points, normalized in bone length units (i.e., independent of the actual skeleton).   From a set of 2D joint and orientation keypoint estimates, the rotations could be determined with a Perspective-n-Point algorithm, such as \cite{Lepetit08epnp:an}.  As our network also learns depth, we use the predictions to reproject XY detections into voxel 3D-space and use these estimates to solve for the transform, which minimizes the least square error from neutral pose.  We found the 3D approach more accurate than PnP and faster. Therefore, we use the method attributed to \cite{Umeyama1991LeastSquaresEO}, which works as follows. Here we define $(\mathbf{x}_{k_i})_{i=1}^6$ as the set of 2 JKPS and 4 OKPS associated with bone $k$. We first compute the centroids of the predictions and the T-pose points ($\mathbf{y}$),

\begin{equation}
\bar{\mathbf{x}}_{k} =\frac{1}{6} \sum_{i=1}^{6} \mathbf{x}_{k_i}, \quad \bar{\mathbf{y}}_{k} =\frac{1}{6} \sum_{i=1}^{6} \mathbf{y}_{k_i}  
\end{equation}
as well as the variance for re-scaling,
\begin{equation}
\sigma^{2}=\frac{1}{6} \sum_{i=1}^{6}\left\| \mathbf{x}_{k_i}-\bar{\mathbf{x}}_{k}\right\|^{2} 
\end{equation}
and the covariance,
\begin{equation}
M_k=\frac{1}{6} \sum_{i=1}^{6}\left(\mathbf{x}_{k_i}-\bar{\mathbf{x}}_{k}\right)^{T}\left(\mathbf{y}_{k_i}-\bar{\mathbf{y}}_{k}\right).
\end{equation}
We then take the Singular Value Decomposition (SVD) of the covariance,
\begin{equation}
M_k=\mathrm{U} \mathrm{S} \mathrm{V}^{\mathrm{T}}
\end{equation}
and compute the rotation of bone $k$ as,
\begin{equation}
R_{k}=U V^{T}
\end{equation}

As JKPS detections are more accurate than OKPS detections, we double the JKPS correspondences' weight in the solution. Using the predicted rotations, we can then trivially infer joint positions to scale from a given skeleton (i.e., bone lengths). Specifically, we use the average bone lengths from the five-subject H3.6m training set.

\subsection{Summary of configurations}

\begin{figure}[h!]
    \centering
    \includegraphics[width=\columnwidth]{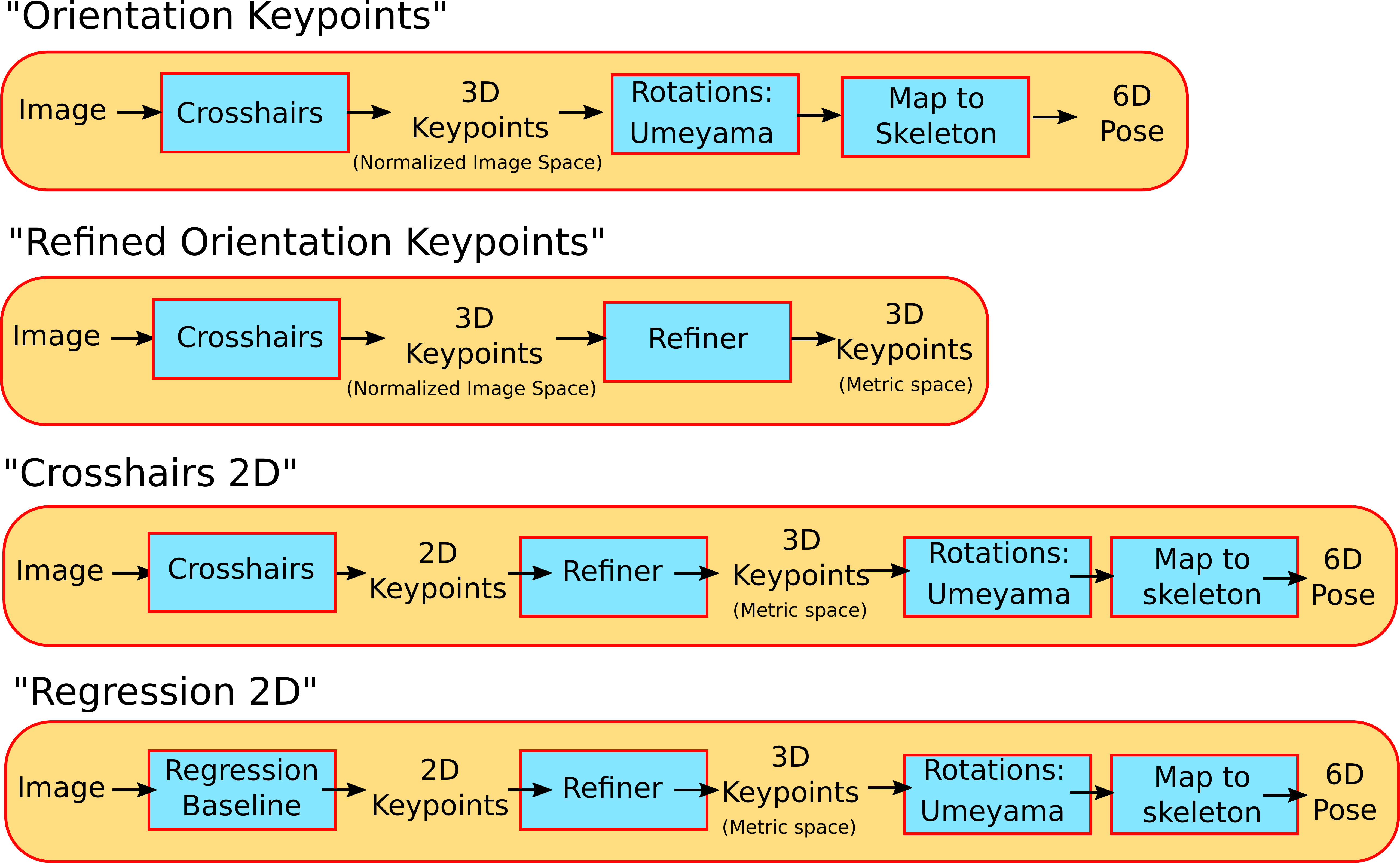}
    \caption{Summary of the pipeline configurations used in this paper. Configuration names correspond to those used in the experiments.}
    \label{fig:configs}
\end{figure}

In the previous sections we defined various detectors options (i.e. 2D/3D and crosshairs/regression) and post-processing methods (i.e. mapping to skeleton or using refiner module). In this paper, we consider four main configurations of these components. Figure \ref{fig:configs} shows the overall pipeline for each configuration. 

All configurations output the 6D pose for each bone of the skeleton. The "Refined orientation keypoints" configuration is special in that it is only used for evaluating the keypoint accuracy when using a refiner for converting from normalized image-space 3D keypoints to metric 3D keypoints.  

\section{Data and Training}
For 3D pose estimation research, the Human3.6m dataset is the most commonly used and includes 3.6 million accurate 3D Human poses acquired by recording the performance of five female and six male subjects, under four different viewpoints, introduced in  \cite{IonescuSminchisescu11} and \cite{h36m_pami}.  Other 3D pose estimation approaches, focused on joint keypoints only, usually train on more diverse datasets and only finetune keypoint locations with Human3.6m, which helps preserve generalization. We found that just training on Human3.6m data was problematic as the model would quickly saturate and struggle to generalize well in validation.  While the set may have many images, there are only five training subjects, and the high parameter models may memorize the specific subjects instead of learning more general rules. We also use the MPII Human Pose dataset \cite{andriluka14cvpr} during training to help with generalization.

\subsection{Losses}
We train our networks to minimize the prediction error over a dataset of poses, where the error is the Mean Per Joint Position Error (MPJPE) which is defined as:

\begin{equation}
    \mathcal{L}_{mpjpe} = \frac{1}{N} \sum_{j=1}^{N} || \mathbf{x}_{j} - \mathbf{x}_{j,gt} ||_2
\end{equation}

where $N = N_o+N_j$. For the crosshairs, we also experimented with a regularizer. This regularizer forces the orientation keypoints at the bone ends towards their centroid.
\begin{equation}
    \mathcal{L}_{cnt} = \sum_k \sum_i ||(\mathbf{x}_{k_i} - \bar{\mathbf{x}}_{k}) - (\mathbf{x}_{k_i,gt} - \bar{\mathbf{x}}_{k,gt})||_2
\end{equation}

We use this to encourage the model to generate a better prediction structure in line with the orientation algorithm used in post-processing.  We expect this to behave similarly to cross-joint loss functions used in the literature, which compare each joint's relative pose to every other joint. However, we use a cross-comparison limited to immediate neighbors.  In our experiments, the accuracy improvement was \textit{de minimis} compared to an identical model trained without the additional loss function.  We did not try this loss function with the simpler regression detector to preserve simplicity.

\subsection{Metrics}

To evaluate the performance of the model, we use MPJPE defined above, along with two angular metrics. Specifically, we define mean average accuracy for rotations as:
\begin{equation}
    MAA = \frac{1}{N}\sum_{j=1}^N 1 - \frac{\theta_{sep}(R_{j,gt},R_{j})}{\pi} 
\end{equation}
Randomly drawn uniformly distributed rotation predictions average approximately 30\% accuracy (with some variation depending on how uniformity is defined for a rotation).

Angular separation maps values to [0, $\pi$] radians and formally is defined as: 
\begin{equation}
    \theta_{sep}(R_m, R_n) = ||log(R_mR_n^T)|| 
\end{equation}
We define MPJAS as the maximum angular separation of points transformed by two rotations:
\begin{equation}
    MPJAS = \frac{1}{N}\sum_{j=1}^N \theta_{sep}(R_{j,gt},R_{j})
    \label{eqn:mpjas}
\end{equation}
Randomly drawn rotation predictions average approximately 2.2 radians error.  MPJAS-15 represents the mean joint comparisons between ground truth and predictions for the 15 free rotations in the typical 17 joint skeleton.

\subsection{Human 3.6m preparation}
Following previous researchers, we extract frames from each video at a downsampled rate of 10HZ (i.e., 1 per 5 frames) and excluded a corrupted sequence for subject 11.  For validation and testing, we tried using all frames, and 1/65 frames (i.e., 1 per 13 at 10hz) as different benchmark papers use different frequencies.  For each image we use a bounding box from ground truth joint data.  We also get comparable but slightly less accurate results when using a fixed pixel bounding box similar to\cite{martinez_2017_3dbaseline}, mainly due to effective resolution loss as the figures have smaller sizes.   Orientation keypoints are calculated, based on the provided angle data and projected into 2D screen coordinates.  For 3x4 aspect ratio resolutions, we preserve scale and crop the width.  For depth detections, we convert depth information into depth equivalent relative to the root.    

\subsection{MPII Human Pose}
MPI-3DHP \cite{mpiiinf3dhp_mono-2017} uses multiple cameras and markerless technology to estimate ground truth data in more varied scenes than Human 3.6m; this capture method is somewhat less accurate than markers as used in Human 3.6m and does not provide joint rotations, and therefore we can only test position accuracy.  The testing is facilitated by a very similar skeleton to Human 3.6m. For the MPII Human Pose dataset \cite{andriluka14cvpr} we bulk adjust annotations for closer consistency with Human3.6m, namely the feet and head.

During the keypoint detector training, we augment the data with random horizontal flipping, rotating by up to +/-30\%, random cropping, positional, and color jitter.  For the lifter/refiner, we use predictions from the first stage and subtract the root location from all 3D keypoints, as is conventional in the literature.  We augment with horizontal flipping and randomly increasing the detector error by 0-100\% vis-\`{a}-vis the ground truth.\newline

\subsection{MPI-3DPW preparation}
MPI-3DPW \cite{vonMarcard2018_3dpw} is a recent dataset that tries to capture full 6D skeletal poses in the wild by using IMUs and a single camera markerless algorithm. The 3DPW dataset uses a different skeleton than the 17 joint skeleton commonly used in the literature. We therefore needed to reconcile the skeleton with Human 3.6m.  While the joint keypoints can generally be matched up, they are placed differently, most noticeably at the hips and the ankles.  For the root hip position, we take the midpoint of LeftUpLeg and RightUpLeg to match Human3.6m.  The other differences still meaningfully impact the MPJPE accuracy: we show substantial improvements in accuracy using our original model from just changing the bone lengths (i.e., narrow hip bone) and even more by fine-tuning to learn the new keypoint locations. 

We also need to calculate rotation accuracy taking into account the additional joints in the kinematic chain (i.e., 3DPW uses a multi-segmented spine) and different orientation conventions (Human 3.6m follows Vicon convention, 3DPW seems to be zero rotation in a T-pose).  To put 3DPW in a common basis, we recalculated the ground truth joint rotations by realigning the rotation matrices (calculated from the provided rotation vectors): we align the Y vector to point from parent to child joint (reversed for lower body), keep the parent annotation Z vector as forward and then orthogonalize and normalize the X and Z vectors.

\subsection{Training regimen}
We begin with transfer learning, using an off-the-shelf Resnet-50 \cite{he2016resnet} backbone used for CPN \cite{chen2018cascaded} as a human keypoint detector on COCO \cite{lin2014microsoft} and then discard the head. For the simple regression detector, we warmup the new head with 1k iterations, and then train the head and layers C4 and C5 of Resnet for 20k iterations at 0.001/0.0001 learning rate, and then another 80k iterations at 0.00025/0.0001.  Each iteration is a 64 sample batch split 75/25 Human3.6m and MPII. For crosshairs we initially train for 40k iterations at 0.001/0.00005 learning rates for the head and backbone, respectively, using L2 loss.  We drop the head learning rate to 0.00025 and train for another 40k iterations.  We then shift the data mix to 75/25 Human36m/MPII and train for another 80k iterations using L1 loss instead.  We use the Adam optimizer \cite{adam_optimizer} and batch normalization \cite{batchnorm}.  

For the second stage refiner, we train for 80 epochs on L2 loss using 0.25 dropout, 0.1 momentum, and the Adam optimizer, starting at a learning rate of 0.001 and 0.98 learning rate gamma.  

\section{Results}

\begin{table}[t!]

\centering
\caption{Rotation results on Human3.6M. We average the 15 free rotations in the typical 17 joint skeleton.  We convert \cite{yoshiyasu_accv_2018} as they report on a different basis.  Low MPJAS (in radians) and high MAA (accuracy) is better. }

\begin{tabular}{lrr}
\hline
& \multicolumn{1}{r}{MPJAS} & \multicolumn{1}{r}{MAA} \\ \hline
Yoshiyasu et. al. \cite{yoshiyasu_accv_2018} ACCV '18 &0.424  &86.5\%   \\
 - pelvis (root) only & 0.226 & 92.8\% \\ \hline
 Orientation Keypoints (ours)\\
 2D detections + PnP & 0.265  & 91.6\%   \\
 3D detections + SVD & 0.213  & 93.2\%   \\
 - pelvis (root) only & 0.145 & 95.4\% \\
 \hline
\end{tabular}%

\label{tab:rotation-results}

\end{table}
\begin{table*}[t!]
\centering
\footnotesize

\caption{Mean Per Joint Position Error (MPJPE) in mm between the ground-truth 3D joints on Human 3.6M for single frame RGB images without depth information. Best results are highlighted in bold.  The ``Orientation keypoints'' results correspond to positions predicted by fitting rotations from the detector to the average training skeleton.  ``Refined orientation kps'' are taken from after the second stage refiner network.  Note that these reference methods use a variety of different training approaches, some include additional data and additional weak supervision.} 

\setlength{\tabcolsep}{2pt}
\begin{tabular}{lrrrrrrrrrrrrrrr|r}
\hline
\textbf{Protocol \#1} & Dir & Disc & Eat & Greet & Phone & Photo & Pose & Punch & Sit & SitD & Smoke & Wait & Walk & WalkD & WalkT & Avg \\ \hline
\multicolumn{1}{l|}{Zhou \etal~\cite{zhou_eccv_16} 6} & 91.8 & 102.4 & 97.0 & 98.8 & 113.3 & 125.2 & 90.0 & 93.8 & 132.2 & 159.0 & 106.9 & 94.4 & 79.0 & 126.0 &  99.0 & 107.3 \\
\multicolumn{1}{l|}{Moreno-Noguer \cite{Moreno-Noguer_2017_CVPR} }
& 66.1 & 61.7 & 84.5 & 73.7 & 65.2 & 67.2 & 60.9 & 67.3 & 103.5 & 74.6 & 92.6 & 69.6 & 71.5 & 78.0 & 73.2 & 74.0 \\ 
\multicolumn{1}{l|}{Pavlakos \etal~\cite{pavlakos2017coarse-to-fine} }
& 67.4 & 71.9 & 66.7 & 69.1 & 72.0 & 77.0 & 65.0 & 68.3 & 83.7 & 96.5 & 71.7 & 65.8 & 59.1 & 74.9 & 63.2 & 71.9 \\
\multicolumn{1}{l|}{Yoshiyasu \etal~\cite{yoshiyasu_accv_2018} }
& 63.3 & 71.6 & 61.4 & 70.4 & 69.9 & 83.2 & 63.1 & 68.8 & 76.8 & 98.9 & 68.2 & 67.5  & 57.7 & 73.7 & 57.1 & 70.0 \\
\multicolumn{1}{l|}{{\color{black}XNect} \cite{XNect} {\color{black}}} 
& 50.2 &61.9 &58.3 &58.2 &68.8 &54.1 &61.5 &76.8 &91.7 &63.4 &74.6 &58.5 &48.3 &65.3 &53.2 &63.0 \\
\multicolumn{1}{l|}{Martinez \etal~\cite{martinez_2017_3dbaseline} } 
& 51.8 & 56.2 & 58.1 & 59.0 & 69.5 & 78.4 & 55.2 & 58.1 & 74.0 & 94.6 & 62.3 & 59.1 & 49.5 & 65.1 & 52.4 & 62.9 \\
\multicolumn{1}{l|}{Yang \etal~\cite{Yang18} }
& 51.5 & 58.9 & 50.4 & 57.0 & 62.1 & 65.4 & 49.8 & 52.7 & 69.2 & 85.2 & 57.4 & 58.4 & 60.1 & \textbf{43.6} & 47.7 & 58.6 \\ 
\multicolumn{1}{l|}{Chen \etal~\cite{Chen_2019_CVPR} }
& 45.9 &53.5 &50.1 &53.2 &61.5 &72.8 &50.7 &49.4 &68.4 &82.1 &58.6 &53.9 &41.1 &57.6 &46.0 &56.9  \\ 
\multicolumn{1}{l|}{Pavlakos \etal~\cite{pavlakos2018ordinal} }
& 48.5 & 54.4 & 54.4 & 52.0 & 59.4 & 65.3 & 49.9 & 52.9 & 65.8 & 71.1 & 56.6 & 52.9 & 44.7 &60.9  & 47.8 & 56.2 \\
\multicolumn{1}{l|}{Luvizon \etal~\cite{Luvizon_2018_CVPR} }
& 49.2 & 51.6 & 47.6 & 50.5 & 51.8& 60.3 & 48.5 & 51.7 & 61.5 & 70.9 & 53.7 & 48.9 & 44.4 & 57.9 & 48.9 & 53.2 \\
\multicolumn{1}{l|}{Pavllo \etal~\cite{DBLP:PavLLO_3Dhumanposeinvideo} } 
& 47.1 & 50.6 & 49.0 & 51.8 & 53.6 & 61.4 & 49.4 & 47.4 &59.3 & 67.4 & 52.4 & 49.5 & 39.5 & 55.3 & 42.7 & 51.8 \\ \hline

\multicolumn{1}{l|}{ {\color{black}Refined jkps only (ours)} } 
&{\color{black}45.1} &{\color{black}50.4} &{\color{black}47.4} &{\color{black}49.1} &{\color{black}54.7} &{\color{black}61.9} &{\color{black}46.3} &{\color{black}45.4} &{\color{black}61.5} &{\color{black}71.4} &{\color{black}51.2} &{\color{black}48.0} &{\color{black}38.6} &{\color{black}52.3} &{\color{black}42.5} &{\color{black}51.1} \\
\multicolumn{1}{l|}{Orientation keypoints (ours)} 
&44.4 &48.9 &42.6 &45.5 &49.8 &50.9 &43.0 &44.4 &56.6 &62.3 &48.3 &44.1 &38.8 &49.5 &42.1 &47.4 \\
\multicolumn{1}{l|}{Refined orientation kps (ours)}
&\textbf{40.7} &\textbf{45.5} &\textbf{39.5} &\textbf{42.3} &\textbf{48.1} &\textbf{49.2} &\textbf{40.3} &\textbf{39.6} &\textbf{56.7} &\textbf{61.3} &\textbf{45.8} &\textbf{41.2} &\textbf{35.3} &46.8 &\textbf{36.8} &\textbf{44.6} \\
\hline \\
\\
\textbf{Protocol \#2 (Procrustes)} & Dir & Disc & Eat & Greet & Phone & Photo & Pose & Punch & Sit & SitD & Smoke & Wait & Walk & WalkD & WalkT & Avg \\ \hline
\multicolumn{1}{l|}{Martinez \etal~\cite{martinez_2017_3dbaseline} } 
& 39.5 & 43.2 & 46.4 & 47.0 & 51.0 & 56.0 & 41.4 & 40.6 & 56.5 & 69.4 & 49.2 & 45.0 & 38.0 & 49.5 & 43.1 & 47.7 \\
\multicolumn{1}{l|}{Chen \etal~\cite{Chen_2019_CVPR} }

& 36.5 &41.0 &40.9 &43.9 &45.6 &53.8 &38.5 &37.3 &53.0 &65.2 &44.6 &40.9 &32.0 &44.3 &38.4 &44.1 \\
\multicolumn{1}{l|}{Pavlakos \etal~\cite{pavlakos2018ordinal} }
& 34.7 & 39.8 & 41.8 & 38.6 & 42.5 & 47.5 & 38.0 & 36.6 & 50.7 & 56.8 & 42.6 & 39.6 & 32.1 & 43.9 & 36.5 & 41.8 \\
\multicolumn{1}{l|}{Pavllo \etal~\cite{DBLP:PavLLO_3Dhumanposeinvideo} } 
& 36.0 & 38.7 & 38.0 & 41.7 & 40.1 & 45.9 & 37.1 & 35.4 & 46.8 & 53.4 & 36.9 & 41.4 &30.3 &43.1  & 34.8 & 40.0 \\ 
\multicolumn{1}{l|}{Yang \etal~\cite{Yang18} }
& \textbf{26.9} & \textbf{30.9} & 36.3 & 39.9 & 43.9 & 47.4 & \textbf{28.8} & \textbf{29.4} & \textbf{36.9} & 58.4 & 41.5 & \textbf{30.5} & 42.5 & \textbf{29.5} & 32.2 & 37.7 \\
\multicolumn{1}{l|}{{\color{black}DaNet~\cite{zhang2020learning} }}
& {\color{black}35.7} & {\color{black}40.4} & {\color{black}39.0} & {\color{black}40.3} & {\color{black}40.5} & {\color{black}47.4} & {\color{black}35.1} & {\color{black}34.9} & {\color{black}45.2} & {\color{black}51.7} & {\color{black}39.6} & {\color{black}37.8} & {\color{black}43.4} & {\color{black}34.4} & {\color{black}39.8} & {\color{black}40.5} \\
\hline
\multicolumn{1}{l|}{{\color{black}Refined jkps only (ours)}} 
&{\color{black}34.2} & {\color{black}38.0} &{\color{black}38.3} &{\color{black}37.8} &{\color{black}39.9} &{\color{black}45.3} &{\color{black}35.5} &{\color{black}33.9} &{\color{black}47.9} &{\color{black}54.2} &{\color{black}40.7} &{\color{black}35.6} &{\color{black}28.9} &{\color{black}39.4} &{\color{black}33.5} &{\color{black}38.9} \\
\multicolumn{1}{l|}{Orientation keypoints (ours)} 
&33.3 &36.3 &34.2 &35.2 &36.3 &38.6 &32.5 &32.8 &42.9 &50.3 &36.5 &33.4 &29.8 &37.2 &32.1 &36.1 \\
\multicolumn{1}{l|}{Refined orientation kps (ours)}
&31.7 &34.5 &\textbf{32.7} &\textbf{33.9} &\textbf{35.4} &\textbf{38.1} &31.5 &30.9 &42.9 &\textbf{49.2} &\textbf{35.8} &32.3 &\textbf{27.6} &35.8 &\textbf{30.5} &\textbf{34.9} \\
\hline \\
\end{tabular}%

\label{tab:mpjpe-table}
\end{table*}

\begin{table*}[t!]
\vspace{-3mm}
\centering
\caption{Results on the 3DHP test set. PCK is percentage correct (within 150mm) of 14 joints, a metric commonly used on this dataset, and MPJPE is again the Mean Per Joint Position Error (mm). PMPJE and PPCK correspond to Protocol 2 where the predictions are further aligned with the ground-truth via a rigid transform using Procrustes before computing MPJPE and PCK, respectively. The datasets on which the models have been trained are shown in brackets. The best results are bolded. Our approach outperforms all the existing methods, even without finetuning on 3DHP itself.
}
\vspace{-3mm}
\begin{tabular}{l|rr|rr|}
& \textbf{MPJPE}  & \textbf{PMPJPE} & \textbf{PCK} & \textbf{PPCK}  \\
& \multicolumn{2}{c|}{\textit{lower is better}} & \multicolumn{2}{c|}{\textit{higher is better}} \\ \hline
\textit{No 3DHP training} & & & & \\
Yang \etal \cite{Yang18} (H3.6m,MPII) &  &  & 69.0 &  \\
Habibe \etal \cite{Habibie_2019_CVPR} (H3.6m) & 127.0 & 92.0 & 69.9 & 82.9 \\
{\color{black}OriNet  \cite{luo2018orinet} (H3.6m)} & {\color{black}-} &{\color{black} -} & {\color{black}71.3} & {\color{black}-} \\
{\color{black}RepNet  \cite{li2020hybrik} (H3.6m, LSP)} & {\color{black}97.8} &{\color{black} -} & {\color{black}82.5} & {\color{black}-} \\
Orientation kps (H3.6m, MPII) & 97.0  &\textbf{ 67.7} & \textbf{81.1} & 93.3  \\ 
Refined orientation kps (H3.6m, MPII) &\textbf{ 94.0}  & 70.7 & 81.7 & \textbf{92.2}  \\ \hline

\textit{Trained with 3DHP} & & & & \\
{\color{black}SPIN} \cite{Kolotouros2019spin} (H3.6m,3DHP,LSP,MPII,COCO) & 105.2 & 67.5 & 76.4 & 92.5 \\ Habibe \etal \cite{Habibie_2019_CVPR} (H3.6m,3DHP) & 90.7 & 65.4 & 81.5 & 91.3 \\
{\color{black}OriNet \cite{luo2018orinet} (MPII, H3.6m, 3DHP)} & {\color{black}89.4} & {\color{black}-} & {\color{black}81.8} & {\color{black}-} \\
{\color{black}XNect} \cite{XNect} (Coco, 3DHP) & 92.4 & - & 82.8 & - \\
{\color{black}MargiPose \cite{nibali20193d} (MPII, H3.6m,3DHP)} & {\color{black}91.3} &{\color{black} -} & {\color{black}85.4} & {\color{black}-} \\
{\color{black}MEVA \cite{nibali20193d} (MPII, H3.6m,3DHP)} & {\color{black}96.4} &{\color{black} 65.4} & {\color{black}-} & {\color{black}-} \\
{\color{black}HybrIK  \cite{li2020hybrik} (H3.6m, 3DHP, COCO)} & {\color{black}91.0} &{\color{black} -} & {\color{black}86.2} & {\color{black}-} \\
Orientation kps fine-tuned (H3.6m,MPII,3DHP)   &\textbf{ 86.1}  &\textbf{ 60.6}  & \textbf{85.8} &

\textbf{94.3 }                                       
\end{tabular}

\label{tab:dhp}
\end{table*}
\begin{table*}[t!]
\centering
\caption{Results on the {\color{black}3}DPW test set. MPJPE is the Mean Per Joint Position Error (in mm), PMPJPE corresponds to MPJPE with the predictions aligned to the ground-truth using a rigid alignment and MPJAS is the Mean Per Joint Angular Separation which measures the angular separation of points transformed by two rotations (see Equation \ref{eqn:mpjas}).  {\color{black} Our method outperforms the current state of the art, Mesh GraphFormer \cite{lin2021mesh} on MPJPE which is the best metric for realworld performance (i.e. without procrustes alignment).}  
}
\vspace{-3mm}
\begin{tabular}{l|rrr|}
\textbf{Method}  & \multicolumn{1}{c}{\textbf{MPJPE}} & \multicolumn{1}{c}{\textbf{PMPJPE}} & \multicolumn{1}{c|}{\textbf{MPJAS}} \\ \hline
SPIN \cite{Kolotouros2019spin} (H3.6m,M3DHP,LSP,MPII,COCO) & 96.9 & 59.2  & -  \\ 
{\color{black} Mesh Graphormer \cite{lin2021mesh} (H3.6m, M3DHP,UP-3D, COCO, MPII, 3DPW)} & {\color{black}74.7} & {\color{black}\textbf{45.6}}  & {\color{black}-}  \\ 
{\color{black} PARE \cite{Kocabas_PARE_2021} (COCO, MPII, LSPET, MPI-INF-3DHP, H3.6m)} & {\color{black}79.1} & {\color{black}46.4}  & {\color{black}-}  \\ 
{\color{black} ROMP \cite{ROMP} (COCO, MPII, LSPET, AICH, MPI-INF-3DHP, H3.6m)} & {\color{black}80.1} & {\color{black}56.8}  & {\color{black}-}  \\ 
{\color{black} HybrIK \cite{li2020hybrik} (COCO, MPI-INF-3DHP, H3.6m)} & {\color{black}80.0} & {\color{black}48.8}  & {\color{black}-}  \\ 
{\color{black} MEVA \cite{nibali20193d} (MPII, H3.6m,3DHP)} & {\color{black}86.9} & {\color{black}54.7}  & {\color{black}-}  \\ 
{\color{black} MeshTransformer \cite{lin2021mesh} (H3.6m, UP-3D, MuCo-3DHP, COCO, MPII, FreiHAND, 3DPW)} & {\color{black}77.1} & {\color{black}47.9}  & {\color{black}-}  \\

\hline
Orientation kps (H3.6m, MPII) & 115.4  & 76.7  & 0.408    \\
Refined Orientation kps (H3.6m, MPII) & 112.4  & 67.6  & -    \\
Orientation kps (H3.6m, MPII, using 3DPW avg bone lengths)     & 90.3   & 66.9  & 0.408 \\
SPIN using Orientation kps in the optimization loop (H3.6m, MPII, 3DPW fine-tuned)   &81.8 &  54.3  & - \\  
Orientation kps fine-tuned (H3.6m, MPII, 3DPW)   & \textbf{70.7} & 50.4   & \textbf{0.302}\\  

\end{tabular}

\label{tab:dpw}
\end{table*}
\begin{table*}[t!]
\footnotesize

\centering
\caption{Ablation study of the lifter showing the effect of different keypoint detectors, and keypoint types. We report MPJPE, PMPJPE and MPJAS when using groundtruth 2D detections, regressed 2D detections, and predictions from our crosshairs architecture (for both 2D and 3D). We show results using only joint keypoints (JKPS) and when augmenting these with orientation keypoints (J+OKPS). The lifter is retrained for each scenario. Our 3D Crosshairs detector performs the best in all cases apart from using ground-truth 2D detections (as expected).
}

\begin{tabular}{l|rr|rrr|rrr|r}
 &
  \multicolumn{2}{c|}{\textbf{Groundtruth 2D}} &
  \multicolumn{3}{c|}{\textbf{Regression 2D (288x384)}} &
  \multicolumn{3}{c|}{\textbf{Crosshairs 2D (288x384)}} &
  \multicolumn{1}{c}{\textbf{Crosshairs 3D}} \\ \hline
Input detections                            & JKPS & J+OKPS & JKPS & JKPS & J+OKPS & JKPS & JKPS & J+OKPS & J+OKPS \\
Output predictions                          & JKPS & J+OKPS & JKPS & J+OKPS & J+OKPS & JKPS & J+OKPS & J+OKPS & J+OKPS \\
Detector 2d err (\% res) & 0\%       & 0\%       & 1.75\%    & 1.75\%    & 1.75\%    & 1.60\%    & 1.60\%    & 1.60\%    & 1.60\%    \\
MPJPE (mm)                    & 48.3      & 33.3      & 57.5      & 57.8      & 53.0      & 50.9      & 50.6      & 47.1      & 44.6      \\
PMPJPE  (mm)                   & 34.9      & 24.9      & 41.0      & 41.0      & 37.6      & 39.5      & 39.4      & 36.0      & 34.9      \\
MPJAS (radians)                             & NA        & 0.150     & NA        & 0.270     & 0.239     & NA        & 0.250     & 0.227     & 0.213    
\end{tabular}

\label{tab:ablation2}
\end{table*}
\begin{table*}[t!]
\centering
\caption{Detector ablation study. We show the effect of different detector architectures (crosshairs and regression), image resolution, number of crosshair layers for XY dimensions, second-stage refinement and training with feet/hands included. For comparability, evaluation averages when training with feet/hands only includes 17 joints and 15 rotations. MPJPE is reported using both Protocol 1 and Protocol 2 (i.e. Procrustes aligned to ground-truth). We see that higher resolution improves the accuracy, and that our crosshairs significantly ourperforms the baseline regression architecture.
} 

\begin{tabular}{lccc|lccc}\hline
 & \textbf{MPJAS} & \multicolumn{2}{c|}{\textbf{MPJPE-17}} &  & \textbf{MPJAS} & \multicolumn{2}{c}{\textbf{MPJPE-17}} \\
 &(15 rot)   & \textbf{P1} & \textbf{P2} &  &(15 rot)   & \textbf{P1} & \textbf{P2}\\ \hline
\multicolumn{4}{l|}{\textit{256x192 Crosshairs (trained on 17j, 15r)} }& \multicolumn{4}{l}{\textit{384x288 Regression (trained on 17j, 15r)} }\\
Detector to skeleton (1-layer XY) & 0.239 & 52.2  & 39.6 & Detector to skeleton  & 0.228 & 52.4 & 40.6  \\
Detector to skeleton (4-layer XY) & 0.235 & 51.9  & 39.1 \\
Refined detections & NA & 49.2  & 38.0 & Refined detections & NA  & 50.0 & 37.6  \\
\hline
\multicolumn{4}{l|}{\textit{384x288 Crosshairs (trained on 17j, 15r)} } & \multicolumn{4}{l}{\textit{384x288 Crosshairs (trained on 21j, 19r)}} \\
Detector to skeleton (1-layer XY) & 0.217 & 48.1 & 37.2 & Detector to skeleton (1-layer XY) & 0.216 & 47.8 & 36.7  \\
Detector to skeleton (4-layer XY) & 0.217 & 48.2 & 37.1 & Detector to skeleton (4-layer XY) & \textbf{0.213} & 47.4 & 36.1\\
Refined detections & NA  & 45.5 & 35.8 & Refined detections & NA  &\textbf{44.6}  &\textbf{34.9}\\
\hline
\\
\end{tabular}

\label{tab:ablation1}
\end{table*}

\subsection{Predicted skeletal rotations}

The literature is mostly devoid of published metrics on rotational prediction accuracy, as even papers purporting to solve for joint rotations instead choose only to show mean positional error.  One exception is \cite{yoshiyasu_accv_2018}, which includes results for Human3.6m.  Their best-stated result equates to 0.424 radians MPJAS-15\footnote{Based on converting their reported metric. As their code is not publicly available, we rely on their brief description in the paper}.  Our best result of 0.213 radians MPJAS-15 is a 48\% improvement on their results and is, as far as we are aware, the {\color{black} state-of-the-art} in predicting skeletal rotations by a considerable margin.

\textbf{Table \ref{tab:rotation-results}} shows summary prediction results for joint rotations on Human3.6m.  We show results with predicted keypoint detection using different strategies to generate rotations based on MPJAS-15, the rotational error across 15 rotations.  First, we use 2D detections of the joint and orientation keypoints to solve with PnP, which achieves 0.265 radians error MPJAS-15 across all actions.  This equates to 91.6\% accuracy and is already a strong result for the first approach.  However, using the full 3D predictions from our detector and SVD we achieve 93.2\% accuracy (0.213 rad).

\subsection{Predicted joint positions}
While localizing joint positions is only the secondary goal of our method, most research focuses on this metric, and we also show meaningful improvements to the {\color{black} state-of-the-art}.  We follow most of the literature in training (S1, S5, S6, S7, S8) / test (S9, S11) split and definitions for Protocol 1 as raw prediction relative to root and Protocol 2 as allowing rigid alignment ('Procrustes') of the overall skeleton.

In Section \ref{sec:post_processing}, we proposed two approaches for estimating the joint positions from the predicted pixel-space values. In Table 2 we can see that, using the first approach, mapping to skeleton, improves the {\color{black} state-of-the-art} MPJPE by 4mm under Protocol 1.  The second approach reduces the mean error by another 3mm, for a total of 7mm.  This is a significant 13.9\% improvement on the previous {\color{black} state-of-the-art} for single frame estimation and better than any reported MPJPE for video analysis as far as we are aware.  Our approach could also be extended to video for potential further improvements. We also establish a new {\color{black} state-of-the-art} under protocol 2.  Again, a simple mapping of detector rotations onto the skeleton improves on the previous {\color{black} state-of-the-art}, and the refinement stage takes the improvement to 3mm.  We also note that the previous {\color{black} state-of-the-art} under this protocol, \cite{Yang18}, uses a GAN to training their detector, an approach that is complementary to our technique.

Qualitatively, compared to other methods, our approach can also ensure a coherent skeleton - no failure cases with elongated limbs.  In \textbf{Table \ref{tab:mpjpe-table}} we report our results compared to various other methods using similar protocols.

\begin{figure}
\centering
  \includegraphics[width=0.45\linewidth]{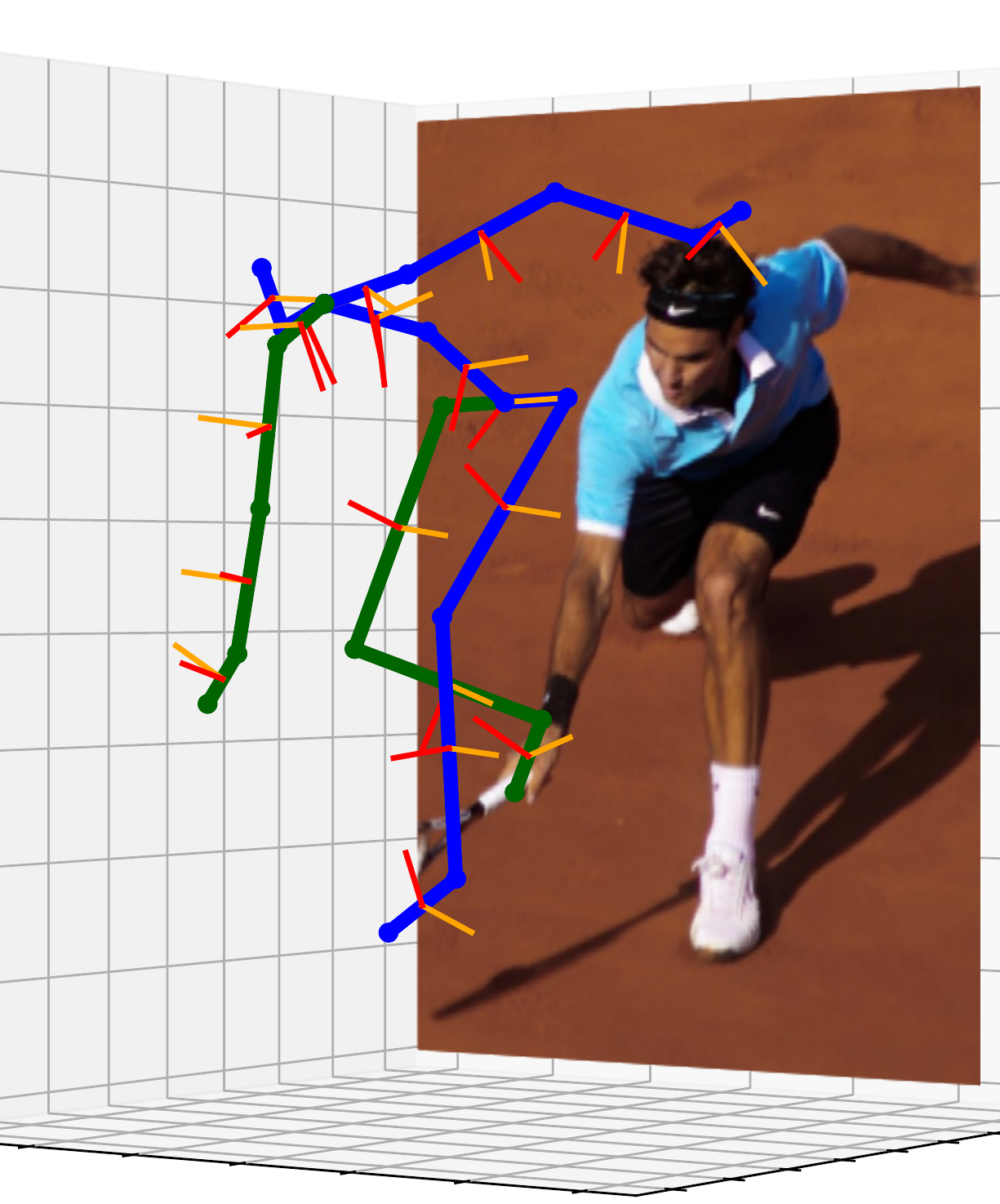}  \includegraphics[width=0.45\linewidth]{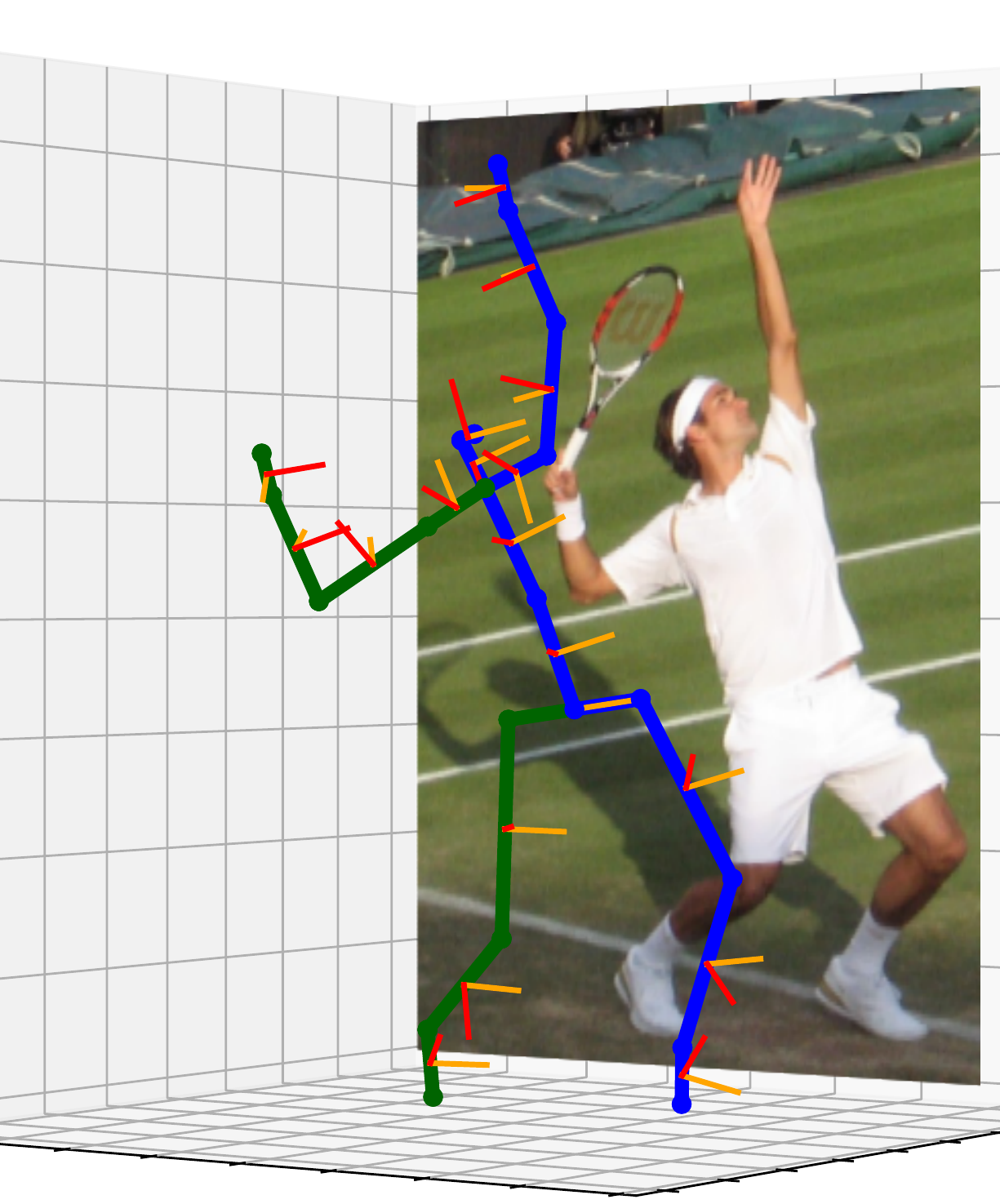}
  \caption{In the wild example of our 6D human pose prediction (Second photo source: \cite{federer2})}
\end{figure}

\subsection{Visualizations}
\begin{figure*}[]
  \includegraphics[width=1.0\textwidth]{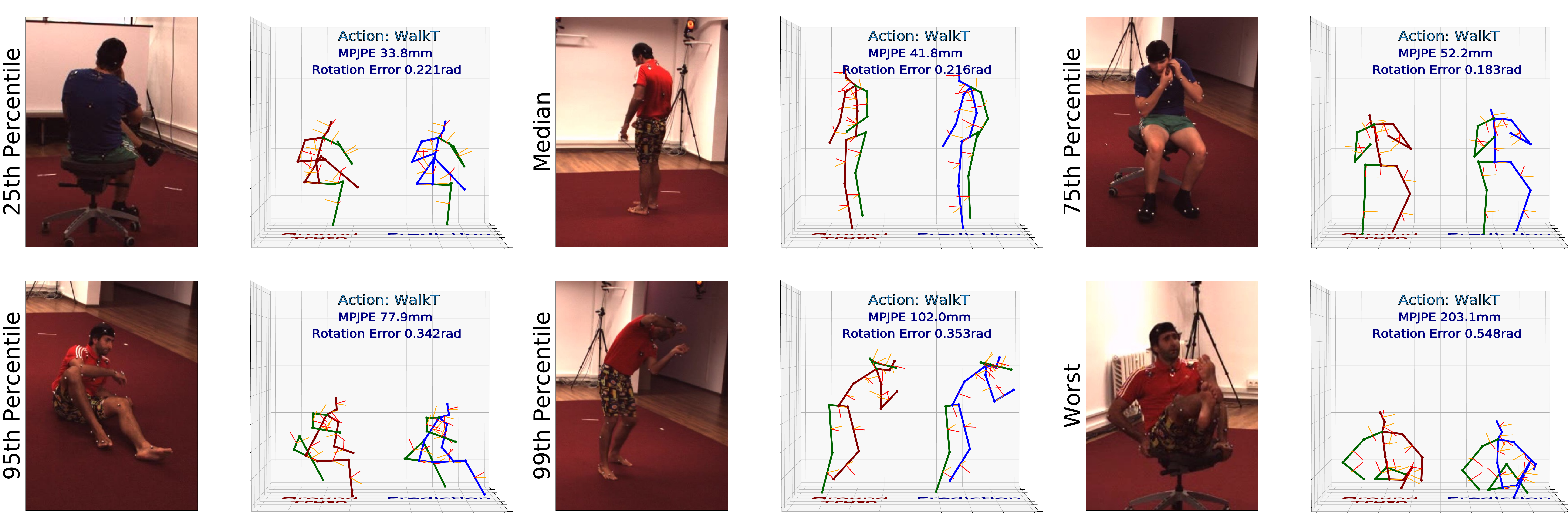}
  \caption{Model predictions versus ground truth. Kinematic rotations are visualized with protruding bone handles - small red lines are joint forward vector, yellow are left.  Samples ranked by Protocol 1 MPJPE on 17 joint skeleton, quoted rotation error is MPJAS-15.  We  visualize feet and hand predictions but do not include in averages}
\label{Fig: visualization}
\end{figure*}

The visualizations provide a more intuitive overview of how our approach performs and demonstrate the rotation information missing from most existing approaches.  To express the full range of detections, we rank the full test set from best (low percentile) to worst (high percentile) based on MPJPE-17.   In \textbf{Figure \ref{Fig: visualization}}, we show key percentile predictions, providing the image with the ground truth skeleton and the predictions plotted side-by-side.  We also show different angles for a better perspective.  To visualize rotations, we plot green and red handles extruding from the bones which show the \textit{forward} and \textit{left} vectors (the bone is always oriented \textit{up}, apart from the legs \textit{down}). 

Most of the predictions are quite accurate, both in terms of overall skeletal form and bone orientation.  The \textit{25th} percentile and median both included heavily occluded limbs yet the model can make highly accurate predictions.  Even the very \textit{worst} result, a pose with the legs heading straight into the camera, is a credible prediction. This example is an outlier, with an error of almost 2x the \textit{99th} percentile.

\subsection{In-the-wild datasets}
We explore how well our approach generalizes to in-the-wild environments by testing on two additional datasets, MPI-3DHP, and MPI-3DPW.  We first use the base model from Human 3.6m unchanged and then show a version fine-tuned with a limited number of iterations at 2.5/1.0 e-4 learning rate (head/backbone) on the relevant training set.  Testing the unchanged model shows how well orientation keypoints generalizes even when trained on a limited dataset.  The fine-tuning allows the model to learn the wider variety of camera angles, distances, and, particularly for 3DPW, learn the significant differences in joint locations.

\textbf{Table \ref{tab:dhp}} shows that our method achieves {\color{black} state-of-the-art} results on 3DHP.  Without training on the 3DHP set, our model is considerably more accurate than the other approaches without 3DHP training and is even competitive with models fully trained on 3DHP.  Our refined model, though, does not perform better than our single-stage model predictions. After 2k iterations of fine-tuning our Human 3.6m model with 3DHP mixed into the batches (48/16 samples per batch 3DHP/H3.6m), the accuracy further improves and is {\color{black} state-of-the-art} across all metrics by a significant margin.  While fine-tuning improves the average metric, the very worst (by MPJPE) failure case worsens further.

\begin{figure*}[]
  \includegraphics[width=1.0\textwidth]{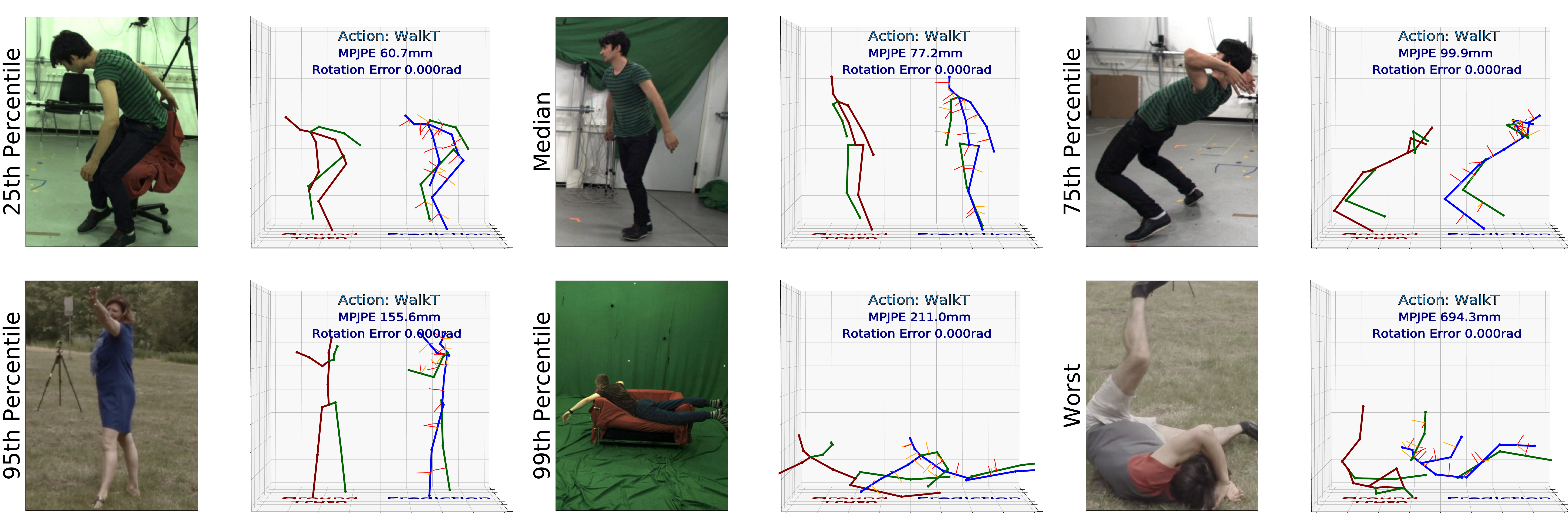}
  \caption{Model predictions for 6 random scenes on the 3DHP dataset (right skeleton) vs ground truth (left skeleton). We show a range of results including the 25th, 50th, 75th, 95th, 99th percentile and as well the example exhibiting the worst error based on MPJPE accuracy. Our approach works well across this entire range. The failure case in the worst example is due to the fact that the detector has confused the left and right limbs. }
\label{Fig: vizDHP}
\end{figure*}

In \textbf{Table \ref{tab:dpw}} we show (i) results of our H3.6m model without further training, (ii) the same model but using bone lengths from the 3DPW training set, (iii) the results after fine-tuning our model for 8k iterations on the 3DPW training/validation set (48/16 samples 3DPW/MPII) and (iv) using the OKPS from our model as an optimisation signal for the mesh fitting loop of the SPIN model as presented in \cite{Kolotouros2019spin}. For (iv) we simply add an L2 loss to the optimization objective that measures the difference between the SMPL joint positions and angles and those estimated from the OKPs. The model generalizes well despite the very different environment and provides {\color{black} state-of-the-art} positional accuracy on this dataset as well. We demonstrate that OKPS can be used to improve other methods, such as SPIN. We also set a benchmark both MPJPE and for accuracy on rotations, 0.302 radians MPJAS, given the noisiness of the provided ground truth, we believe this is a solid result and demonstrates good generalization of our Human 3.6m results. We do note that the method employed to estimate ground-truth for 3DPW has 0.208 radians of error in their test environment, quantitatively comparable to our results on Human 3.6m, and is likely noisier on the provided in-the-wild data. The test set has significant and visible annotation errors when merely re-projecting the ground truth. Nevertheless, the dataset provides an opportunity to benchmark rotations in a more challenging setting.

\begin{figure*}[]
  \includegraphics[width=1.0\textwidth]{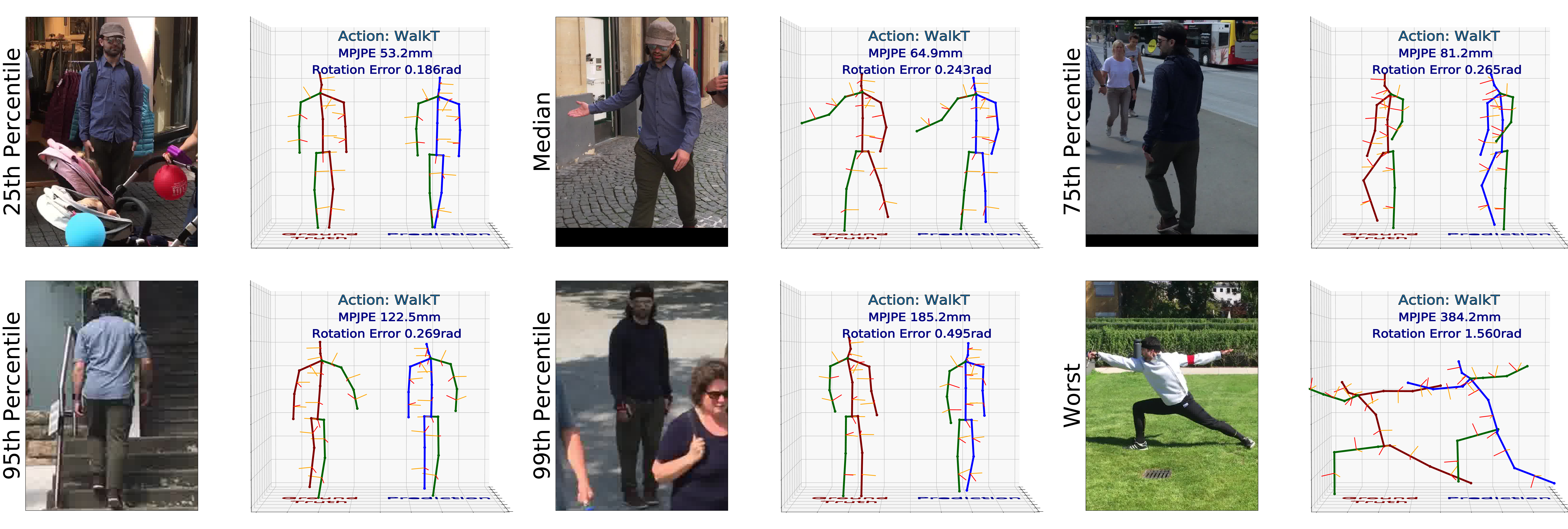}
  \caption{Model predictions on the 3DPW dataset (right skeleton) vs ground truth (left skeleton). Again we show a range of results including the 25th, 50th, 75th, 95th, 99th percentile and as well the example exhibiting the worst error based on MPJPE accuracy. Overall our approach works well across all the examples, with the failure case again attributed to a swapped limb detection. }
\label{Fig: vizDPW}

\end{figure*}

We visualize both of these datasets in \textbf{Figures \ref{Fig: vizDHP}} and \textbf{\ref{Fig: vizDPW}}, again showing a range of results based on MPJPE accuracy.  The failure cases are illustrative.  For 3DHP, the detector gets the entire body reversed for the quantitatively worst prediction.  For 3DPW, the worst prediction is for a lunging fencer, and the model rotates the torso circa 120 degrees to reverse the arms.  A few frames earlier, before the fencer is at full extension, the model does not make this mistake; incorporating video analysis may resolve these kinds of ambiguities.  In other cases, such as 75th and 95th percentile examples, the skeletal scaling is the culprit for MPJPE as the rotations are quite accurate.

We also show two visualizations of in-the-wild images with Table \ref{tab:rotation-results}.  Despite the absence of sports images in the Human3.6m training set, the model has generalized well to a completely different context. \newline

\subsection{Detector network and ablation}
The accuracy of our approach, like that of \cite{martinez_2017_3dbaseline,DBLP:PavLLO_3Dhumanposeinvideo} and others, is impacted by the accuracy of the keypoint detector.  The {\color{black} state-of-the-art} in detectors is rapidly evolving, and different models offer different accuracy versus computational and training complexity tradeoffs.  Higher-resolution input can also improve detector accuracy.  We focus this paper on the benefits of using orientation keypoints to solve for 6D human rotations - our method can be applied to different detectors.  We therefore explore the most critical aspects of detector impact here in several ways.     

\textbf{Ground truth analysis}.  Following \cite{martinez_2017_3dbaseline} and \cite{DBLP:PavLLO_3Dhumanposeinvideo}, we use ground truth analysis to provide a baseline as it separates the issue of detection quality, a moving target given rapid advances in the field, and focuses on the information content of orientation keypoints.  As shown in \textbf{Table \ref{tab:ablation2}}, lifting orientation keypoints with ground truth data improves MPJPE by 15mm.  This clearly shows the added value of orientation keypoints for human pose estimation irrespective of the detector employed.

\textbf{A simpler detector}.  We also show, in \textbf{Table \ref{tab:ablation2}} and \textbf{Table \ref{tab:ablation1}}, the results of our simple detector, a regression head with Resnet50, on accuracy using both 2D and 3D versions.  The detector is less accurate than Crosshairs but still delivers {\color{black} state-of-the-art} results as OKPS improves MPJPE accuracy by 4.5mm and 3.4mm.

\begin{figure}[h!]
  \centering
  \includegraphics[width=\linewidth]{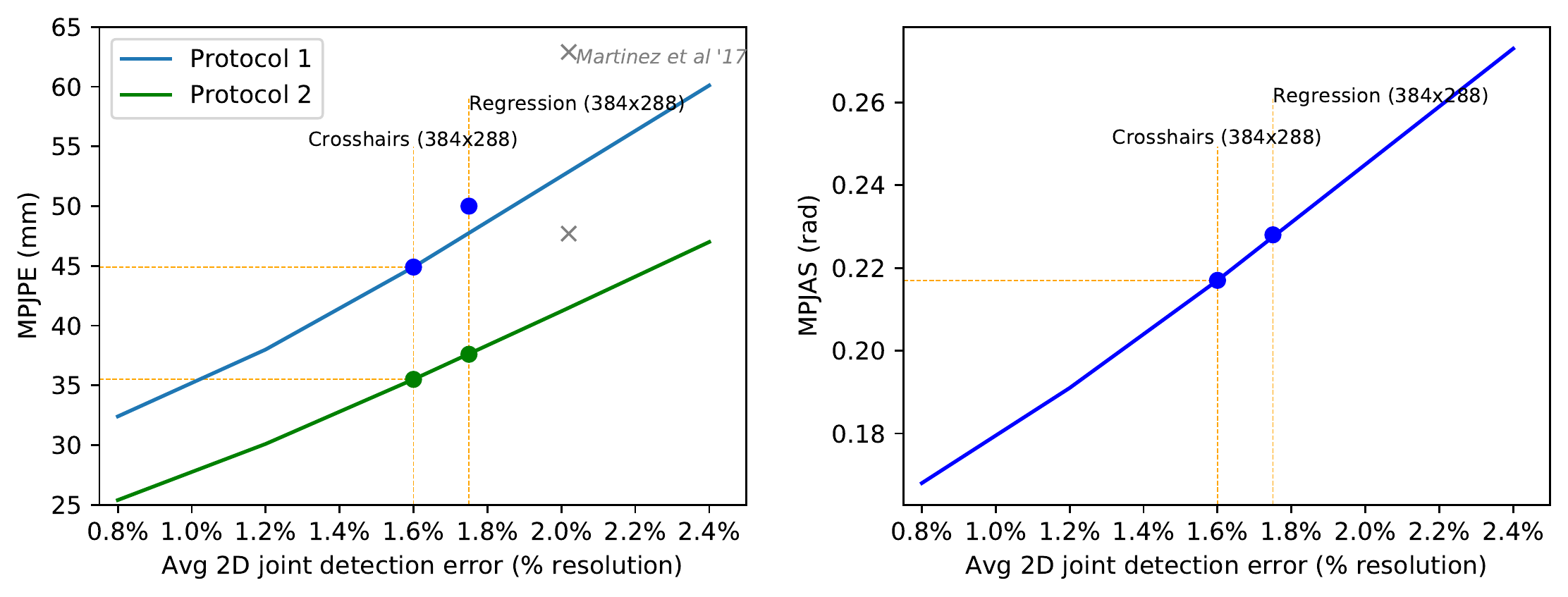}
  \caption{Orientation keypoints sensitivity to detector accuracy.  We plot both detectors we used and the fine-tuned stacked hourglass detections from \cite{martinez_2017_3dbaseline}.  The lines are generated from refining detections with errors scaled.  Note: detection errors are overstated as 3 actions for S9 are misannotated in source data as highlighted in \cite{LearnableTriangulation}}
\label{fig: det_sensitivity}
\end{figure}


\textbf{Detector accuracy simulation.}  We can also use the refiner stage to analyze results by directly scaling the detection errors over the ground truth.  We show the impact of detector accuracy on MPJAS and MPJPE in \textbf{Figure \ref{fig: det_sensitivity}}.  This gives a sense of potential improvements from further detector advances as well as potential degradation from a more challenging dataset, and we believe more informative than applying Gaussian noise as it preserves the correlation structure of the detector errors.  Using the detections of  \cite{martinez_2017_3dbaseline}, we plot their stacked hourglass detector as a reference - at their level of accuracy, our method reduces MPJPE by 15\%, while providing full skeletal rotations.  Figure \ref{fig: det_sensitivity} also shows that errors in 6D are correlated to errors of the detector.

\textbf{Lifting OKPS only from JKPS}.  As we show in \textbf{Table \ref{tab:ablation2}}, we can also use a lifter model to predict both joint and orientation keypoints from only the 2D JKPS: this is sufficient to make an excellent prediction, achieving 0.250 radians MPJAS.  This further demonstrates that OKPS is a useful representation to predict rotations.  Of course, predicting OKPS from the image benefits from additional clues and is more accurate. 

\textbf{Further ablation}
We decompose the benefits of different resolutions, multi-layer crosshairs, and the extra refiner network in Table \ref{tab:ablation1}.  
Each earlier layer increases the computation cost of flattening as the kernel size increases.  While the intermediate supervision has a clear benefit, inference on the validation set of multiple crosshairs only offers a modest detection benefit at the higher resolution.  As early layer crosshairs are computationally more expensive, these could be dropped for deployment, leaving a very lightweight head (only 11\% incremental multiply-adds over the Resnet-50 backbone).  Similarly, the lower resolution version offers competitive accuracy at half the computational cost.   

\subsection{Comparison to mesh-based approaches}

Approaches that regress dense mesh correspondences have become an important paradigm for pose estimation, and therefore we provide a numerical comparison to a state-of-the-art method of this type. 

In particular, we compare our approach to DaNet \cite{zhang2020learning}, which embodies the dense mesh regression paradigm. This method predicts the dense correspondences between 2D pixels and 3D vertices with an HRNet-W48 which has similar capacity to our backbone and then uses this as an intermediate representation to estimate rotations of body joints via a GCN. The results are reported in Table \ref{tab:mpjpe-table}. As shown in the table, even though DaNet has a similar capacity, all configurations of our method outperform this approach on H3.6m. 

We also compare to HybrIK \cite{li2020hybrik} which is a recent SMPL-based approach. From Table \ref{tab:dpw} we can see that our approach significantly outperforms HybrIK on 3DPW on all the reported metrics.  Additionally, to demonstrate the complementary nature of our research, we used our OKPS predictions as an extra optimisation parameter for the SMPL-based SPIN approach from \cite{Kolotouros2019spin} and improved upon the original results by a substantial 15.1 MPJPE.

\section{Conclusion}

We have proposed a novel approach for human pose estimation that makes use of orientation keypoints to parameterize bone orientations and have demonstrated that our method significantly improves upon the {\color{black} state-of-the-art}.  We can accurately predict skeletal rotations from a single RGB camera image while accurately localizing joints in 3D.  Our technique is simple and straightforward to apply, and we believe it can become a vital part of the human pose estimation pipeline. 

Our approach offers many opportunities for extension. Further advances in detection methods should improve our results.  Exploiting other datasets should improve accuracy, and our method could be used with weak supervision strategies and GANs.  Extending our work to video is a natural step as temporal information can help resolve ambiguities.  Training to generate a customized skeleton (i.e. bone lengths) could also improve precision.  Finally, our post-processing is differentiable and could be directly incorporated for training.


\end{document}